\pdfoutput=1
\documentclass[11pt]{article}

\usepackage{EMNLP2023}
\usepackage{times}
\usepackage{latexsym}
\usepackage[T1]{fontenc}
\usepackage[utf8]{inputenc}
\usepackage{microtype}
\usepackage{inconsolata}
\usepackage{graphicx}
\usepackage{booktabs}
\usepackage{multirow}
\usepackage{color}
\usepackage{makecell}
\usepackage{subcaption}
\usepackage{array}

\title{Investigating the Nature of Disagreements on Mid-Scale Ratings:\\A Case Study on the Abstractness–Concreteness Continuum}

\author{Urban Knuple\v{s}$^1$ \and Diego Frassinelli$^2$ \and Sabine Schulte im Walde$^1$ \\
  $^1$Institute for Natural Language Processing, University of Stuttgart \\
  $^2$Department of Linguistics, University of Konstanz \\
  \texttt{\{urban.knuples, schulte\}@ims.uni-stuttgart.de} \\
  \texttt{diego.frassinelli@uni-konstanz.de} \\}

\begin{document}

\maketitle

\begin{abstract}
Humans tend to strongly agree on ratings on a scale for extreme cases (e.g., a \textsc{cat} is judged as very concrete), but judgements on mid-scale words exhibit more disagreement. Yet, collected rating norms are heavily exploited across disciplines. Our study focuses on concreteness ratings and (i) implements correlations and supervised classification to identify salient multi-modal characteristics of mid-scale words, and (ii) applies a hard clustering to identify patterns of systematic disagreement across raters. Our results suggest to either fine-tune or filter mid-scale target words before utilising them.
\end{abstract}

\section{Motivation}

Across disciplines, researchers have collected and exploited human judgements on semantic variables such as concreteness, compositionality, emotional valence, and plausibility. Traditionally, those judgements are collected as a degree on a continuum between extremes. While humans tend to strongly agree on their ratings for extremes (e.g., a \textsc{cat} is typically judged as extremely concrete; \textsc{glory} as extremely abstract; the compound \textsc{crocodile tears} as extremely non-compositional; \textsc{war} as extremely negative), we find considerable disagreement regarding human mid-range ratings, i.e., judging about semi-concreteness, semi-compositionality, semi-negativity. Presumably, conceptual \textit{semi-}properties are not easily graspable, thus generating stronger disagreement among raters. Nevertheless, the collected norms are heavily exploited in state-of-the-art computational approaches, where the respective knowledge represents a crucial task-related component (such as concreteness information for figurative language detection, and emotional valence for sentiment analysis).

\begin{figure*}[ht!]
    \centering
    \includegraphics[width=0.9\linewidth]{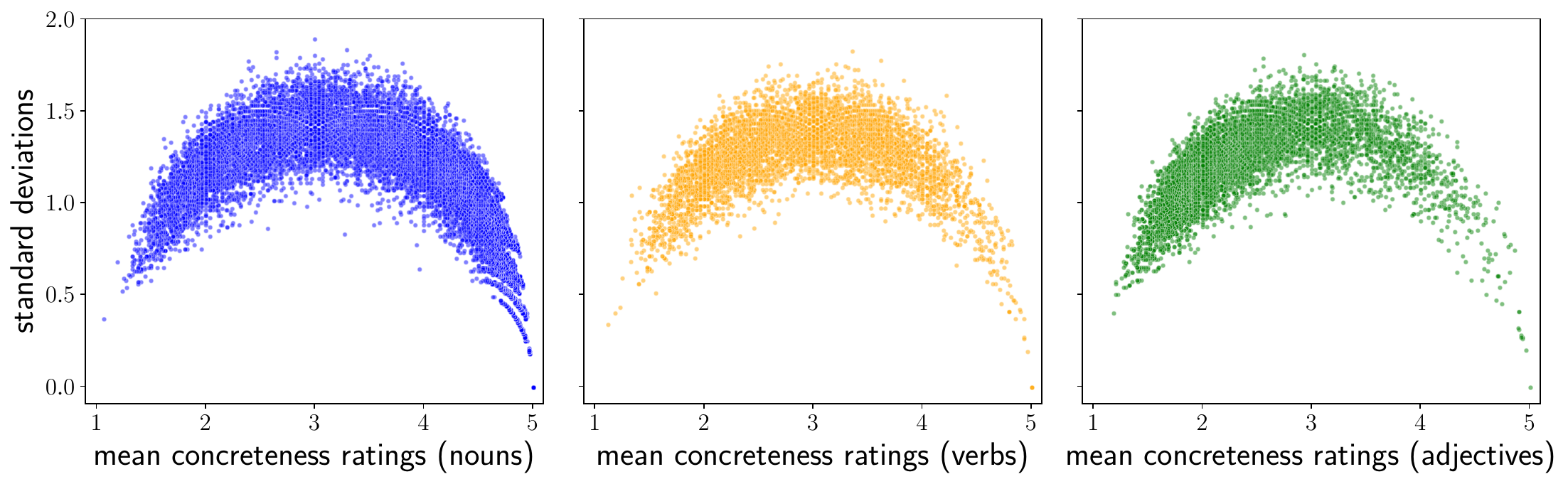}
    \caption{Croissant plots -- Mean concreteness scores and standard deviations of ratings in \citet{BrysbaertEtAl:14}.}
    \label{fig:conc-rating-distributions}
\end{figure*}

The current study provides a series of analyses on human mid-scale ratings, while focusing on the most prominent collection of concreteness ratings for English words \citep{BrysbaertEtAl:14}, henceforth \textit{Brysbaert norms}. As basis for the Brysbaert norms, humans were asked to judge the concreteness (in contrast to abstractness) of English words on a 5-point rating scale from $1$ (abstract) to $5$ (concrete) regarding how strongly the participants thought the meanings of the targets can(not) be experienced directly through their five senses. Figure~\ref{fig:conc-rating-distributions} illustrates the distribution of the mean concreteness ratings and standard deviations (SDs) across 25 raters and for the three word classes of nouns, verbs, and adjectives. These \textit{croissant}\footnote{We use this term due to the shape of the distribution plots.} plots for ratings on a scale can exhibit ``only a finite number of possible combinations of means and standard deviations'' \citep{Pollock:18}: humans tend to agree on the extremes ($\rightarrow$ low SD) and to disagree on intermediate \textit{semi-}values ($\rightarrow$ high SD).

In a first set of experiments, we analyse multi-modal characteristics of the concreteness of target nouns in the Brysbaert norms (we provide additional materials for verbs and adjectives in the Appendix): perception strength for specific senses (auditory, gustatory, haptic, olfactory, visual), emotional dimensions (valence, affect, dominance), lexical properties (frequency, ambiguity) and association types as indicators of meaning diversity. We start with a holistic perspective via correlations between targets' concreteness and their characteristics, and then zoom into differences for words with mid-scale vs.\ extremely concrete/abstract mean concreteness ratings, by applying supervised classification and feature analyses. In a second set of experiments, we hypothesise that mid-scale ratings are due to different combinations of individual human judgements across the scale. We thus rely on the original per-participant ratings (i.e., 25 ratings per target) and apply exploratory cluster analyses to identify patterns of disagreement between the individual raters of targets with mid-scale ratings.

Our contributions in this paper are two-fold. (i)~We identify a range of target word characteristics that overall correlate with their degrees of concreteness ratings in different directions, and more specifically differ for mid-scale and extremely concrete or abstract target words. (ii)~We identify a range of systematic disagreement patterns that clearly differ across target words with mid-scale mean ratings, thus pointing out fine-grained differences in judgements on semi-perception and suggesting to either filter or fine-tune mid-scale target words before utilising them in computational approaches.

In the remainder of this paper, we introduce previous related work (Section~\ref{sec:related}) and our concreteness targets (Section~\ref{sec:targets}); we then report our analyses regarding general and mid-scale target characteristics (Section~\ref{sec:analyses-characteristics}) and mid-scale disagreement patterns (Section~\ref{sec:analyses-disagreement}).

\vspace{+2mm}
\section{Related Work}
\label{sec:related}

Collecting human judgements on a rating scale is a popular means of constructing concept-specific datasets across languages, research disciplines and (computational) linguistics tasks. Prominent example tasks and collections targeting semantic variables include compositionality ratings for compound--constituent relatedness \citep[i.a.]{ReddyEtAl:11a, SchulteImWaldeEtAl:16a, CordeiroEtAl:19, GagneEtAl:19, GuentherEtAl:20}, affect ratings such as valence, arousal, dominance, emotion \citep[i.a.]{Kanske/Kotz:10, Koeper/SchulteImWalde:16a, Mohammad:18}, plausibility ratings \citep[i.a.]{WangEtAl:18, Eichel/SchulteImWalde:23}, and concreteness ratings \citep[i.a.]{Spreen/Schulz:66, PaivioEtAl:68, AlgarabelEtAl:88, DellaRosaEtAl:10, BrysbaertEtAl:14, Koeper/SchulteImWalde:16a, BoninEtAl:18, MurakiEtAl:22}.

As a main motivation for collecting general conceptional ratings on a scale, \citet{Keuleers/Balota:15} state that there is ``no reason for words to be rated for every single experiment''. Still, researchers across disciplines have pointed out problematic aspects of rating norms, because their reliability is unclear, especially when ratings have been collected via crowdsourcing or extrapolation \citep{Keuleers/Balota:15, ManderaEtAl:15}. \citet{Pollock:18} describes the typical shape of ratings on a scale, pointing out that the mid-range concepts are the least agreed upon, and that the interpretation of the corresponding ratings conflates \textit{semi-}properties and genuine disagreements. A mid-scale score in concreteness could thus refer to an average \textit{semi-}perception (whatever this means), or to a specific \textit{semi-}sense, such as vision, haptics, etc., as well as to disagreement about perceptual strength, or a combination of the above. Furthermore, many conceptual ratings have been collected by presenting the word in isolation without reference to the respective word class and out of context. For example, the Brysbaert norms rely on isolated target presentation, and part-of-speech information was added post-hoc from the SUBTLEX-US corpus \citep{BrysbaertEtAl:12}. \citet{MurakiEtAl:22} used the same setup as \citet{BrysbaertEtAl:14} but for multiword expressions, in which case part-of-speech ambiguity did not arise, but the targets were also presented out of context.

Despite these problems, ratings on a scale still remain the major strategy to collect human judgements on degrees of semantic variables, while alternatives such as best-worst scaling are available \citep{Kiritchenko/Mohammad:17, AbdallaEtAl:23}. The resulting norms are heavily exploited in state-of-the-art computational approaches; e.g., emotion and concreteness norms represent a crucial component in systems to detect figurative language usage \citep{TurneyEtAl:11, TsvetkovEtAl:14, Koeper/SchulteImWalde:16b, MohammadEtAl:16b, AedmaaEtAl:18, Koeper/SchulteImWalde:18, MaudslayEtAl:20}. The current study encourages researchers to distinguish between degrees of (dis)agreement of such norms, and to identify a meaningful way of exploitation, in particular for mid-scale ratings.

\section{Concreteness Targets and Ratings}
\label{sec:targets}

\begin{figure*}[ht!]
    \centering
    \includegraphics[width=\textwidth]{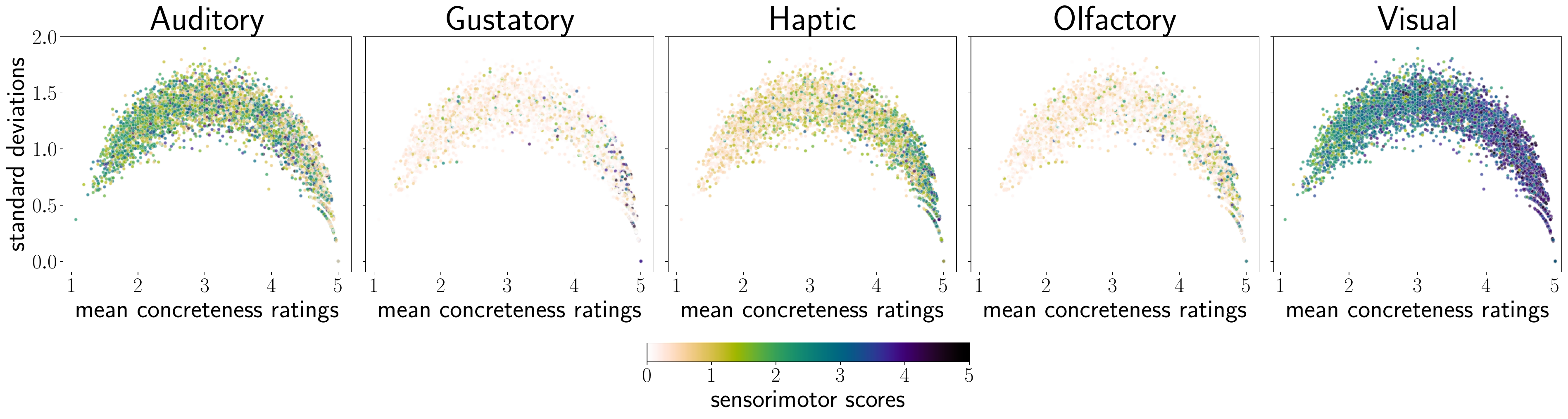}
    \caption{Mean noun ratings and standard deviations overlaid with the respective sense perception scores.}
    \label{fig:sensorimotor_nouns}
\end{figure*}

As materials for our experiments, we utilise the concreteness norms collected by \citet{BrysbaertEtAl:14}, including approximately $40,000$ English target words.\footnote{We disregard any two-word expressions.} The resource contains individual ratings by 25 participants on a 5-point scale ranging from 1 (abstract) to 5 (concrete), mean ratings and standard deviations. No context or part-of-speech (POS)  were given; in a post-processing step, \citet{BrysbaertEtAl:12} added POS and frequency information from the SUBTLEX-US corpus.

We followed a further post-processing step suggested by \citet{SchulteImWalde/Frassinelli:22}, who assigned the most frequently occurring POS tag and frequency information to the target words using the ENCOW web corpus \cite{Schaefer/Bildhauer:12, Schaefer:15}, and then reduced the targets to a less ambiguous and less low-frequent subset by discarding words for which (i) the predominant POS did not represent at least $95\%$ of all POS occurrences; (ii) the newly assigned ENCOW POS tag was not identical to the SUBTLEX-US POS tag, or (iii) for which the ENCOW target frequency was lower than $10,000$. Our subset includes $5,448$ nouns, $1,280$ verbs and $2,205$ adjectives, and is publicly available.\footnote{\url{http://www.ims.uni-stuttgart.de/data/mid-scale}}

\section{Target Words: Characteristics}
\label{sec:analyses-characteristics}

In our first set of experiments we analyse multi-modal characteristics of our concreteness targets. After introducing these characteristics (Section~\ref{sec:characteristics}), we start out with a holistic perspective by quantifying statistical relationships between degrees of concreteness and our selection of target characteristics (Section~\ref{sec:characteristics-holistic}). We then zoom into differences in characteristics between mid-scale target words and extremely concrete/abstract target words, by applying a classifier that determines separability based on characteristics (Section~\ref{sec:characteristics-mid-extreme}).

\subsection{Characteristics and Resources}
\label{sec:characteristics}

\vspace{+2mm}
\paragraph{Sense Perception}
Given that the original concreteness ratings in the Brysbaert norms rely on the raters' perceptions across senses, the most intimately connected set of characteristics explores the relationships between concreteness ratings and the five senses that were used in the task definition by \citet{BrysbaertEtAl:14} when collecting judgements for the concreteness norms. While \citeauthor{BrysbaertEtAl:14} did not ask for a reference to specific senses rather than a general strength of sense perception, \citet{LynottEtAl:20} collected judgements on specific senses (auditory, gustatory, haptic, olfactory, and visual) for the same targets as \citeauthor{BrysbaertEtAl:14}, using a scale from $0$ to $5$. 

\paragraph{Emotion Dimensions}
Abstract words are considered to be more emotionally valenced than concrete words \citep{KoustaEtAl:11, ViglioccoEtAl:14, Pollock:18}. We thus explore emotion dimensions of our target words by using the NRC VAD Lexicon \cite{Mohammad:18}\footnote{\url{https://saifmohammad.com/WebPages/nrc-vad.html}} with ratings on valence, arousal, and dominance for over $20,000$ commonly used English words. The ratings were obtained by asking participants to judge the VAD strength of words using a best-worst scaling method. For each emotion dimension, the scores range from $0$ (lowest VAD) to $1$ (highest VAD).

\paragraph{Frequency and Ambiguity}
Frequency and ambiguity represent two standard dimensions influencing language processing and comprehension \citep[i.a.]{Ellis:02,BaayenEtAl:16}. For frequency information, we rely on the target frequencies extracted from the ENCOW web corpus (see Section~\ref{sec:targets}), containing $\approx$10 billion words. In order to distinguish between degrees of ambiguity of the targets, we rely on WordNet \cite{Miller/Fellbaum:91, Fellbaum:98}, a standard lexical semantic taxonomy for English word senses developed at Princeton University. WordNet organises words into classes of synonyms (\textit{synsets}) connected by lexical and conceptual semantic relations. We looked up the number of noun and verb (but not adjective) target senses in WordNet version 3.0 and then used these WordNet ambiguity values if in the range $[1; 6]$; targets with more than six senses in WordNet we assigned to a joint additional category.

\paragraph{Free Word Associations} 

Previous work suggested that free associations to abstract words differ from free associations to concrete words in terms of the number of types, thus pointing towards differences in conceptual semantic diversity. At the same time, associations to concrete words have been found weaker and more symmetric than for abstract words \citep{Crutch/Warrington:10, HillEtAl:14b}. The \textit{Small World Of Words Project} SWOW \cite{deDeyneEtAl:19}\footnote{\url{https://smallworldofwords.org/}} provides large databases with free word associations across languages; for English, SWOW-EN includes more than $12,000$ cue words with responses from over $90,000$ participants. The associations were gathered from 2011--2018 by asking English speakers through crowd-sourcing to produce the first three response words that came to mind when presented with a cue word. We rely on SNOW-EN associations as indicators of diversity regarding our target words. Next to using only the first response \textsc{R1}, we aggregated the first two responses into a set \textsc{R12}, and all three responses into a set \textsc{R123} to decrease sparsity, while accepting a minor association chain effect\footnote{According to the association chain effect, the $n$th association response is supposedly associated to the ($n$-1)th association response rather than being associated to the target word; this effect might contaminate later association responses.} \citep{McEvoy/Nelson:82, SchulteImWalde/Melinger:08}. We measured the diversity of responses by counting the number of types (i.e., the number of distinct associations that were produced across participants) in \textsc{R1}, \textsc{R12}, and \textsc{R123}, and normalised by the respective total numbers of response tokens.

\begin{table}[ht!]
\centering
\vspace{+2mm}
\begin{tabular}{lrrr}
\toprule

& \multicolumn{1}{c}{N} & \multicolumn{1}{c}{V} & \multicolumn{1}{c}{A} \\
\midrule
\midrule
Targets in our subsets & 5,448 & 1,280 & 2,205 \\
\midrule
Sense perception & 5,440 & 1,280 & 2,202 \\
\midrule
Emotion & 5,012 & 1,104 & 1,987 \\
\midrule
Frequency & 5,448 & 1,280 & 2,205 \\
Ambiguity\footnote{We did not include adjectives.} & 5,400 & 1,277 & -- \\
\midrule
Diversity: associations & 3,501 & 780 & 1,255 \\
\bottomrule
\end{tabular}

\caption{Coverage of target characteristics.}
\vspace{-1mm}
\label{tab:target-number}
\end{table}

\paragraph{Word Classes and Resource Coverage}

Table~\ref{tab:target-number} provides an overview of how many of our targets are covered by the various resources across word classes. Note that from now on the main body of this paper will focus on nouns, and additionally we will refer to supporting evidence or differences regarding verb and adjective analyses in the text and in the Appendix.

\subsection{Holistic Perspective}
\label{sec:characteristics-holistic}

\begin{figure*}[!htbp]
    \centering
    \includegraphics[width=\linewidth]{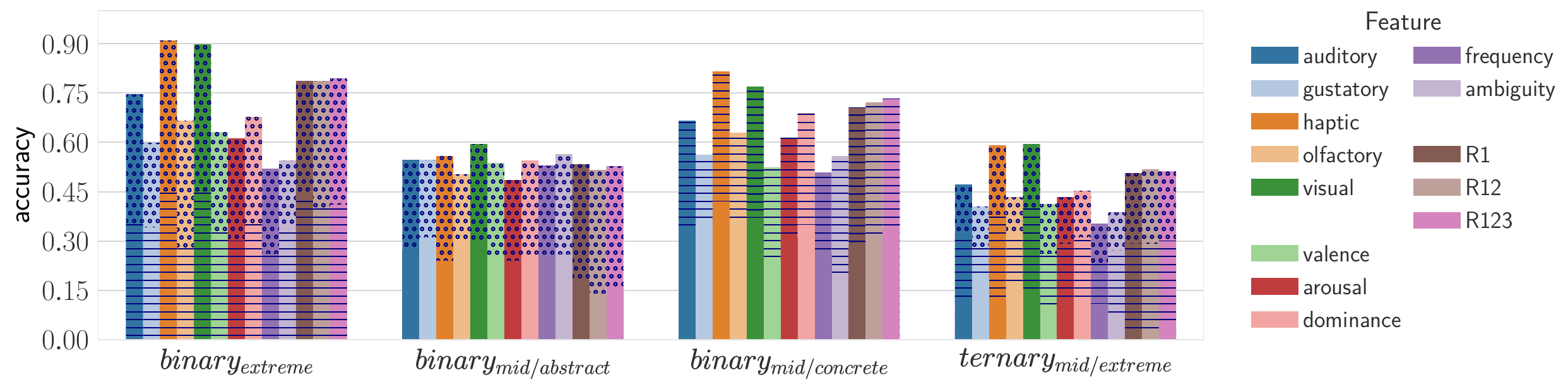}
    \caption{Results of classifications across characteristics and mid-scale/extreme experiments. The dotted and horizontal line patterns indicate the amount of abstract and concrete nouns correctly classified.}
    \label{fig:mid-scale-mean-classification-results-nouns}
    \vspace{+2mm}
\end{figure*}

Figure~\ref{fig:sensorimotor_nouns} visualises the relationships between mean noun concreteness ratings and standard deviations as introduced in Figure~\ref{fig:conc-rating-distributions}, in combination with heat maps indicating the rating strengths of auditory, gustatory, haptic, olfactory and visual perception (left to right).\footnote{Plots for further characteristics are in Appendix~\ref{sec:app-plots-croissant-noun}.} Targets missing in a resource are plotted in grey. We can clearly observe an overall dominance of the visual perception (also see Table~\ref{tab:modalities} in Appendix~\ref{sec:app-dominance-perception} for perception across senses), and that the strength of perception varies in different ways across the concreteness rating scale. 

Table~\ref{tab:correlations2} informs us that visual, haptic, and olfactory sense perception (positively), as well as auditory (negatively), correlate with the noun concreteness scores. Regarding further target characteristics, the table reports a negative correlation with emotion regarding affect and dominance, as well as negative correlations with concept diversity regarding association types. The lexical characteristics do not show any correlations with concreteness.

\begin{table}[ht!]
\centering
\begin{tabular}{llr}
\toprule

\multicolumn{2}{c}{Target characteristics} & \multicolumn{1}{c}{$\rho$} \\
\midrule

\multirow{5}{*}{Sense perception} & Auditory & -0.28$^{*}$ \\
                 & Gustatory & 0.01$^{\textcolor{white}{*}}$ \\
                 & Haptic & 0.58$^{*}$ \\
                 & Olfactory & 0.29$^{*}$ \\
                 & Visual & 0.61$^{*}$ \\
\midrule
\multirow{3}{*}{Emotion} & Valence & -0.01$^{\textcolor{white}{*}}$ \\
        & Affect & -0.28$^{*}$ \\
        & Dominance & -0.32$^{*}$ \\
\midrule
\multirow{2}{*}{Lexicon}          & Frequency & -0.00$^{\textcolor{white}{*}}$ \\
                 & Ambiguity & -0.11$^{*}$ \\
\midrule
\multirow{3}{*}{Diversity: associations} & R1 & -0.33$^{*}$ \\
                        & R12 & -0.41$^{*}$ \\
                        & R123 & -0.43$^{*}$ \\

\bottomrule
\end{tabular}

\caption{Spearman's rank-order correlation coefficient $\rho$ for the statistical relationships between degrees of concreteness and strengths of target noun characteristics; significance level is $p<0.001$.}
\label{tab:correlations2}
\vspace{-2mm}
\end{table}

We thus conclude that overall the concreteness ratings of our target nouns\footnote{See Tables~\ref{tab:correlationsV}--\ref{tab:correlationsA} in Appendix~\ref{sec:app-correlation-verb-adj} for verbs and adjectives.} correlate to different degrees -- and differing in negative vs.\ positive directions -- with specific senses and also with further characteristics previously attributed to abstract vs.\ concrete concepts. This is our starting point for analysing whether any of these characteristics is particularly different for mid-scale target words and might have influenced their concreteness ratings.

\begin{table}[ht!]
    \centering
    \begin{tabular}{lcc}
    \toprule
    Classification variants & Baseline & \multicolumn{1}{c}{Accuracy} \\
    \midrule
    \textit{binary$_{extremes}$} & 0.50 & 0.98 \\
    \midrule
    \textit{binary$_{mid/abstract}$} & 0.50 & 0.75 \\
    \textit{binary$_{mid/concrete}$} & 0.50 & 0.93 \\
    \midrule
    \textit{ternary$_{mid/extremes}$} & 0.33 & 0.79 \\
    \bottomrule
    \end{tabular}
    \caption{Overall classification results (accuracy).}
    \label{tab:mid-scale-mean-classification-results-nouns}
\vspace{-2mm}
\end{table}

\subsection{Mid-Scale Peculiarities}
\label{sec:characteristics-mid-extreme}

We now investigate more specifically genuine characteristics of words that received mid-scale ratings, by zooming into differences in characteristics of mid-scale in contrast to extremely concrete/abstract target words, to maximise contrasts. For this, we created three sets of 500 nouns each: the 500 most abstract nouns, the 500 most extreme nouns, and the 500 nouns with mean ratings closest to the rating-scale mean of $3$ (with 250 nouns with mean $\le3$ and 250 nouns with mean $>3$).\footnote{We created several variants of mid-scale definitions, but given that neither modelling results nor insights differ strongly, we provide the variants in Appendix~\ref{sec:app-mid-scale-def-results}.} 
\begin{figure*}[ht!]
    \centering
        \includegraphics[width=.45\linewidth]{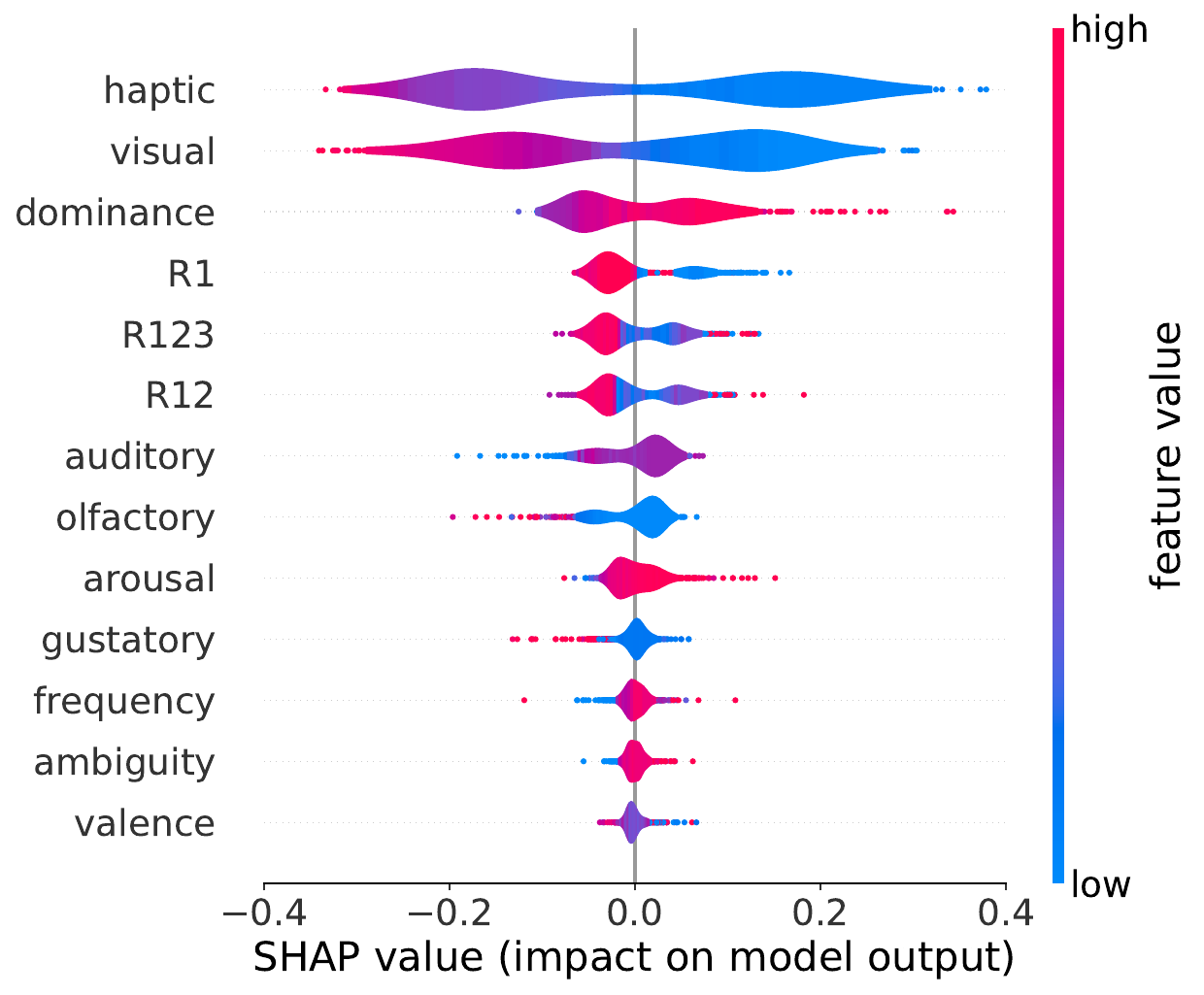} \hspace{1cm}
    \includegraphics[width=.45\linewidth]{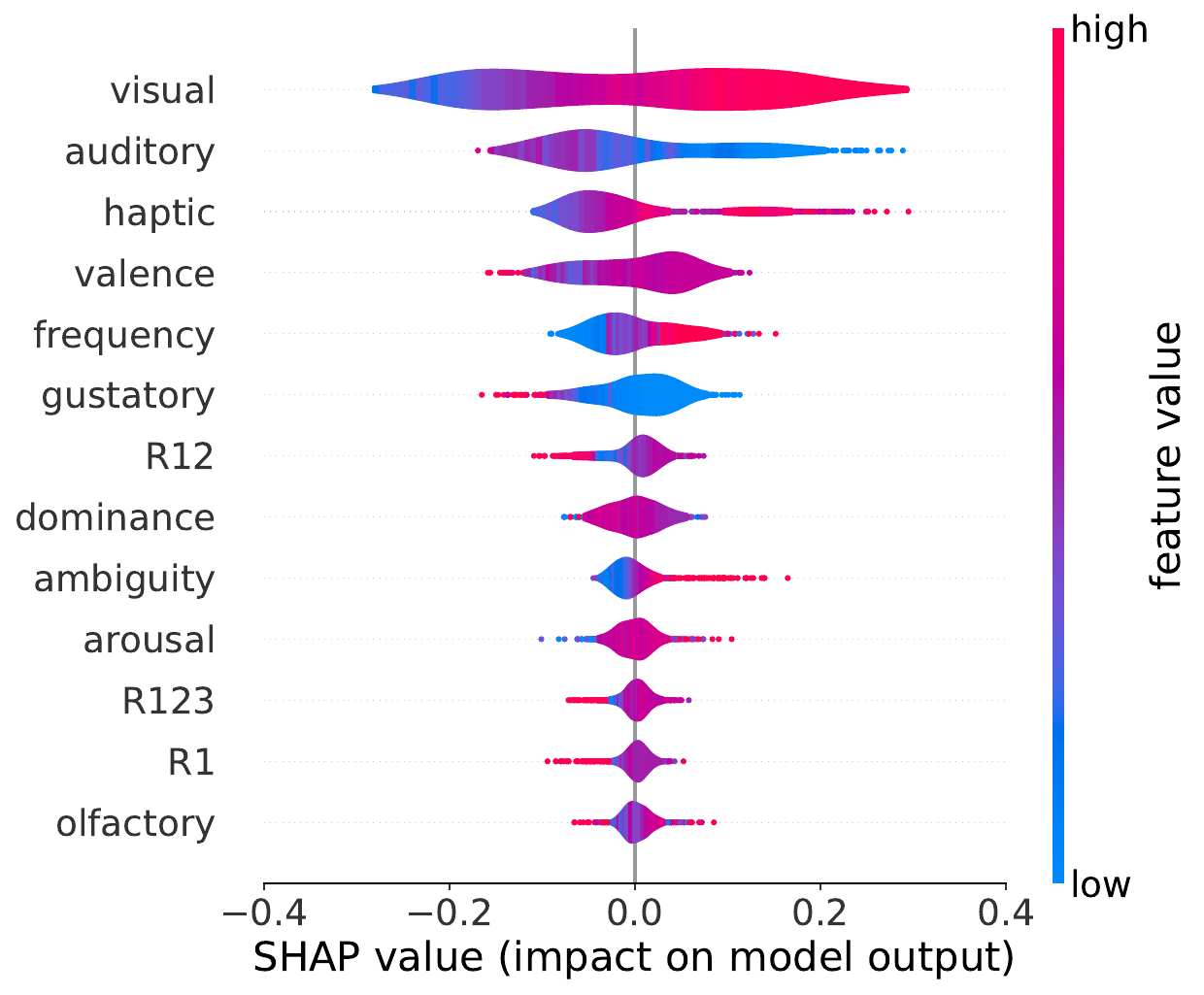}
    \caption{SHAP values -- Importance of each feature for the output of the  \textit{binary$_{mid/concrete}$} model (on the left) and the \textit{binary$_{mid/abstract}$} model (on the right). Extreme nouns are coded as negative, mid-scale nouns as positive.}
    \label{fig:SHAP}
\end{figure*}
We then applied a Random Forest classifier and defined the following classification variants: a \textit{ternary$_{mid/extremes}$} condition where the  classifier had to distinguish between the two extreme sets of 500 concrete and abstract targets from the mid-scale; \textit{binary$_{mid/abstract}$} and \textit{binary$_{mid/concrete}$} conditions to zoom into the individual mid-scale vs.\ extreme differences. As a control condition providing an upper bound for our classifiers, we included \textit{binary$_{extremes}$} where we classify only the extreme target sets with stronger differences between the two classes, while disregarding the mid-scale sets. The respective baselines are $50\%$ for the binary classifications and $33\%$ for the ternary classification.

The classifier used as features those target characteristics described and analysed in Section~\ref{sec:characteristics-holistic}, separately and combined, in order to identify the characteristics that differ for mid-scale words in contrast to clearly abstract or concrete words. If a target word lacks a feature for a specific variable, we assigned $0$ as the respective feature value. We applied 10-fold cross-validation and report the average accuracy score. The classification results using all the features at the same time are shown in Table~\ref{tab:mid-scale-mean-classification-results-nouns}. Figure~\ref{fig:mid-scale-mean-classification-results-nouns} shows the results per feature type. As expected, the \textit{binary$_{extremes}$} classifications show the best results, with auditory, haptic, and visual sense perception as well as association diversity representing the strongest characteristics, in accordance with their overall correlation strengths in Section~\ref{sec:characteristics-holistic}. The \textit{ternary$_{mid/extremes}$} results look like a miniature version of the \textit{binary$_{extremes}$} results with regard to accuracy across feature types, only on a lower scale (given the extra class). The results for the \textit{binary$_{mid/abstract}$} and \textit{binary$_{mid/concrete}$} conditions are lower than for \textit{binary$_{extremes}$}, as predicted, because the contrasts on the concreteness scale are less strong. Also, we observe an interesting difference between the two conditions: targets with mid-scale ratings are distinguished better from targets with extremely concrete  in comparison to extremely abstract ratings ($\rightarrow$~higher accuracy); at the same time, feature contributions in \textit{binary$_{mid/concrete}$} are similar to those in \textit{binary$_{extremes}$} and \textit{ternary$_{mid/extremes}$}, while their contributions in \textit{binary$_{mid/abstract}$} are more uniform. 

To further understand the differences between these two conditions, we inspected the contribution of each feature to the models' output using Shapley Additive Explanations (SHAP; \citealp{LundbergLee2017}). Figure~\ref{fig:SHAP} shows the importance --~as the magnitude of change~-- of each variable in predicting the concreteness scores of concrete (left plot) and abstract (right plot) nouns vs.\ mid-scale nouns. The colours of the violin plots indicate the values of the features. For the \textit{binary$_{mid/concrete}$} model, the three most important features for the classification are haptic, visual, and dominance, in that order. Conversely, for the \textit{binary$_{mid/abstract}$} model, the most important features are visual, auditory, and haptic. Notably, visual and haptic features emerge as the most informative in both cases. Associations, instead, show a relatively small contribution to the performance of the classifier when together with other feature types (as opposed to Figure~\ref{fig:mid-scale-mean-classification-results-nouns}).

An analysis of the colour-coded information (i.e., the value of each feature) supports our previous evidence. In the left plot in Figure~\ref{fig:SHAP}, we can see a clear distinction between concrete nouns that are characterised by high (magenta) visual and haptic values, and mid-concreteness nouns characterised by low (blue) visual and haptic values. Conversely, in the right plot in Figure~\ref{fig:SHAP} the visual and haptic nature of abstract versus mid-scale nouns exhibits less pronounced differences with magenta colour associated both with mid-scale (positive) and abstract (negative) nouns.

We thus infer from our classification experiments that mid-scale target nouns are more easily distinguishable from extremely concrete in comparison to extremely abstract targets, with regard to our set of features. In the next section, we will investigate why this is the case.

\section{Mid-Scale Disagreement Patterns}
\label{sec:analyses-disagreement}
\begin{figure*}[ht!]
    \centering
    \includegraphics[width=.7\linewidth]{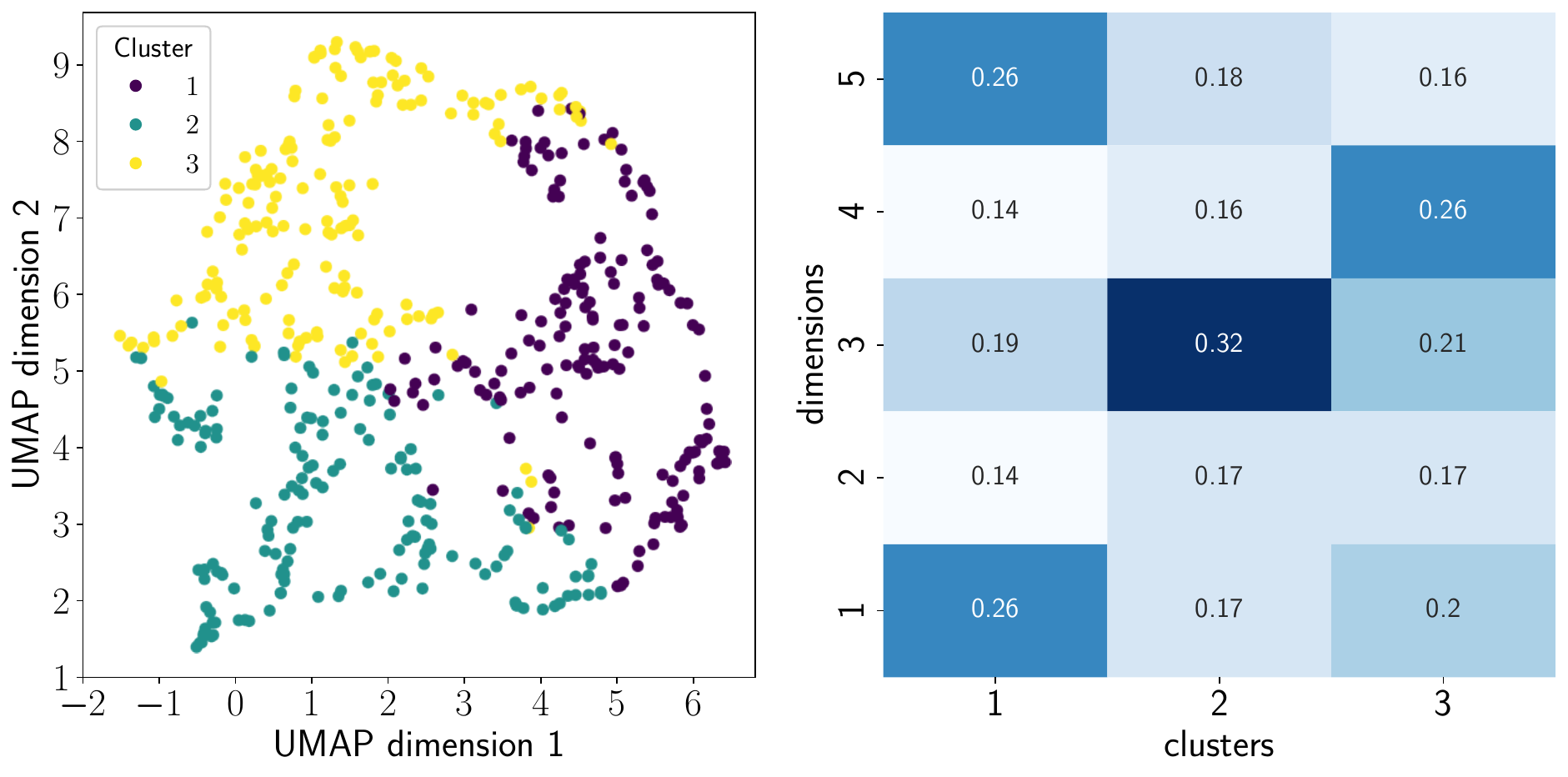}
    \caption{$k$-Means clustering ($k=3$) of 500 mid-scale nouns based on original individual per-participant rating distributions. Cluster sizes are 170, 163, and 167. The heatmap shows the rating distributions of the centroid vectors.}
    \label{fig:clustering_mean_nouns}
\end{figure*}

In our final analyses, we zoom into the numerical characteristics of mid-scale mean ratings. If there was substantial agreement behind the \textit{semi}-perception of a mid-scale target (i.e., if all human raters had provided a rating of $3$ or similar on the scale of $1$ to $5$), then we would see a standard deviation around $0$ in the croissant plots in Figure~\ref{fig:conc-rating-distributions}. We however observe rather high standard deviations for targets with mean ratings of $\approx$$3$, thus indicating considerable disagreement across raters. The question we are asking is how these disagreements were triggered. We hypothesise that raters might have been influenced differently by their individual perceptions of target characteristics, and that we therefore find several patterns of disagreement across the mid-scale target words.

For this exploration of disagreement patterns, we make use of the original per-participant ratings in \citet{BrysbaertEtAl:14}, and applied a standard $k$-means hard clustering approach to automatically assign the 500 mid-scale nouns to $k=3$ clusters. As representations for the targets, we used $5$-dimensional vectors with relative frequencies per rating categories ${1, 2, 3, 4, 5}$, based on the original individual ratings, e.g., the vector for the noun \textit{discussion} is $\vec{v} = \langle 0.15, 0.07, 0.48, 0.15, 0.15 \rangle$, because 15\% of the raters provided ratings of $1, 4$ or $5$, while 7\% judged it as~$2$, and 48\% judged it as~$3$.

Figure~\ref{fig:clustering_mean_nouns} presents two perspectives on the resulting clusters with rather homogeneous cluster sizes $170, 163, 167$. On the left,\footnote{We used UMAP (Uniform Manifold Approximation and Projection) for down-scaling our distributions to two dimensions \citep{McInnesEtAl:18}.} we can see that the three clusters are clearly separated, with relatively small overlapping areas, thus indicating that the underlying cluster features (i.e., the rating distributions) clearly differ. This is confirmed by the plot on the right, which shows the individual rating distributions ($y$-axis) of the three cluster centroids $1$--$3$ ($x$-axis). The heatmap exhibits rather different patterns: in cluster $1$, we find the strongest disagreements among raters, where each of the two extreme rating scores ($1$ and $5$) were chosen by 26\%, the mid-score by 19\%, and the remaining scores are equally distributed over ratings $2$ and $4$ (14\% each); in cluster $2$, 32\% of the raters judged the respective target nouns as~$3$ because they were completely undecided or they consciously chose a mid-scale \textit{semi-}perception score, while the other raters decided for $1,2,4,5$ with almost identical proportions (16--18\%);
finally, in cluster~$3$ we find a more uniform rating distribution, while a score of~$4$ was given by most of the raters (26\%). Table~\ref{tab:ex-cluster-targets} provides a few example targets for each of the three clusters, together with their rating distributions.

\begin{table}[h!]
    \centering
    \begin{tabular}{c|l|l}
    \toprule
    C & Target & Distribution \\
    \midrule
    \multirow{3}{*}{1}
    & \textit{definition} & $\langle 0.32,0.11,0.14,0.11,0.32 \rangle$ \\
    & \textit{hero} & $\langle 0.22,0.11,0.26,0.19,0.22\rangle$ \\
    & \textit{percentage} & $\langle 0.40,0.03,0.10,0.20,0.27 \rangle$ \\
    \midrule
    \multirow{3}{*}{2}
    & \textit{coward} & $\langle 0.17,0.20,0.30,0.20,0.13 \rangle$ \\
    & \textit{discussion} & $\langle 0.15, 0.07, 0.48, 0.15, 0.15 \rangle$ \\
    & \textit{labor} & $\langle 0.16,0.12,0.40,0.12,0.20 \rangle$ \\
    \midrule
    \multirow{3}{*}{3}
    & \textit{booster} & $\langle 0.32,0.07,0.14,0.29,0.18 \rangle$ \\
    & \textit{election} & $\langle 0.20,0.10,0.23,0.27,0.20 \rangle$ \\
    & \textit{hour} & $\langle 0.23,0.07,0.23,0.30,0.17 \rangle$ \\
    \bottomrule
    \end{tabular}
    \caption{Examples of rating distributions for noun target words across clusters C.}
    \label{tab:ex-cluster-targets}
\end{table}

Overall, Figure~\ref{fig:clustering_mean_nouns} thus provides very strong evidence in favour of our hypothesis that a mid-scale mean rating conflates rather different patterns of disagreements across human ratings. Figures~\ref{fig:clustering_mean_verbs} and~\ref{fig:clustering_mean_adjectives} in Appendix~\ref{sec:app-clusters-verb-adj} provide the respective plots for verbs and adjectives, where we find similar patterns of disagreement.

\section{Discussion \& Conclusion}

We started out with the well-known observation that humans tend to strongly agree on ratings on a scale for extreme cases, but that judgements on mid-scale words exhibit more disagreement. This observation is well-described by the croissant-like shape of mean rating scores in relation to their standard deviations (cf. Figure~\ref{fig:conc-rating-distributions}). While individual studies have pointed out problems with such ratings on a scale (e.g., \citet{Kiritchenko/Mohammad:17, Pollock:18}) and also provided alternative settings (e.g., \citet{Kiritchenko/Mohammad:17, AbdallaEtAl:23}), the scale-based norms are heavily exploited across disciplines, including state-of-the-art computational approaches.

In the current study, we first asked whether words with mid-scale concreteness ratings potentially exhibit specific characteristics that genuinely distinguish them from clearly concrete and clearly abstract words. The corresponding classification experiments and feature analyses demonstrated that mid-scale targets were indeed distinguishable from extreme targets with regard to a subset of the senses which were used as criteria for the concreteness--abstractness distinction (mainly visual and haptic), and also with regard to emotional dimensions and meaning diversity (implemented on the basis of association types). In this first set of experiments mid-scale targets therefore established themselves as genuine intermediate concepts. We also saw, however, that mid-scale nouns are more easily distinguishable from extremely concrete in comparison to extremely abstract nouns, and this asymmetry flips with regard to verbs and adjectives, presumably because their underlying rating distributions exhibit different skews (cf. the croissant plots in Figure~\ref{fig:conc-rating-distributions} and the different mid-scale ranges in Figure~\ref{fig:mid-scale-ranges} in Appendix~\ref{sec:app-mid-scale-def-results}). So overall, words with mid-scale mean ratings represent rather genuine intermediate concepts regarding our implementations of features and analyses.

In the second part of our study, we investigated whether mid-scale ratings are generally agreed upon across raters, or whether raters disagreed regarding their \textit{semi-}perception. Relying on explorative cluster analyses using the original per-participant rating distributions, we found clusters with obviously very different centroids. From this, we induce that a mid-scale rating mean of $\approx$3 conflates rather different yet systematic kinds of disagreements. This observation is in line with the mathematically-based observations by \citet{Pollock:18} that ``there is only a finite number of possible combinations of means and standard deviations'', and at the same time it clearly demonstrated that mid-scale ratings indeed differ regarding their underlying rating combinations. So, on the one hand, our cluster analyses confirm a so-far rather theoretically-driven observation; on the other hand, we raise the question of whether and how this observation should influence the utilisation of ratings on a scale. We suggest two alternative routes: (i)~either filter the norm targets and only keep those targets that are clearly attributable to one extreme, or (ii)~fine-tune the mid-scale norm targets with regard to inherent disagreement patterns, because the set of mid-scale targets is itself rather inhomogeneous but nevertheless provides valuable information regarding specific differences in human perception.

Last but not least we would like to point out that inherent disagreements among human annotators are obviously not restricted to our particular focus on mid-scale ratings but represent a common issue under discussion across annotation tasks. In the past decade the field has moved from considering disagreements as pure noise towards zooming into disagreements in order to distinguish between noise and subjectivity, and to effectively exploit the  value of disagreements in language modelling, see
\citet{Alm:11} and \citet{UmaEtAl:21} for a prominent opinion paper and a prominent survey, respectively.

Our analyses and insights should be interpreted in the same vein: we attribute disagreements on concreteness mid-scale ratings to genuine intermediate concepts (see above) and suggest to take a fine-grained approach when utilising them in language modelling tasks and applications.

\section*{Limitations}

Our study is targeting ratings on a scale but currently restricted to a selection of target properties and a specific case study on concreteness. Future work will explore additional target properties that might influence concreteness mid-scale ratings (such as the mass-count distinction and register) as well as characteristics of ratings on a scale in further collections and other languages than English.

\section*{Ethics Statement}

For our study, we used and cited publicly available datasets and libraries.  The resources do not contain any information that uniquely identifies individuals. Our research does not raise any immediate ethical concerns.

\section*{Acknowledgements}

This research was supported by the Ad Futura Scholarship (305. JR) from the Public Scholarship, Development, Disability and Maintenance Fund of the Republic of Slovenia (Urban Knuple\v{s}), and by the DFG Research Grant SCHU 2580/4-1 (\textit{MUDCAT -- Multimodal Dimensions and Computational Applications of Abstractness}). We also thank the reviewers for suggesting additional perspectives regarding our analyses and interpretations.

\bibliographystyle{acl_natbib}
\bibliography{conll-submission}

\clearpage
\appendix
\label{sec:appendix}

\setcounter{page}{1}

\onecolumn

\section{Dominance of Perception across Targets}
\label{sec:app-dominance-perception}

Table~\ref{tab:modalities} shows how many of our target words (nouns, verbs, adjectives, overall) were perceived pre-dominantly by any of the human senses auditory, gustatory, haptic, olfactory, visual, according to the analyses by \citet{LynottEtAl:20}.

\begin{table*}[h]
\centering
\vspace{+8mm}
\fontsize{10}{12}\selectfont
\setlength{\tabcolsep}{3pt}
\begin{tabular}{lrrrrrr}
\toprule
{} &  Auditory &  Gustatory &  Haptic &  Olfactory &  Visual &  Total \\
&           &            &         &            &         &        \\
\midrule
N  &        610 &        199 &     102 &         38 &    4,491 &   5,440 \\
V   &        269 &          8 &      27 &          4 &      972 &   1,280 \\
A &        341 &         31 &      64 &          7 &    1,759 &   2,202 \\
\midrule
all &      1,220 &        238 &     193 &         49 &    7,222 &   8,922 \\
\bottomrule
\end{tabular}
\caption{Distribution of dominant perceptual modalities of our target words, based on \citet{LynottEtAl:20}.}
\label{tab:modalities}
\end{table*}

\clearpage
\section{Visualisations of Rating Characteristics for Nouns\footnote{The corresponding visualisations of rating characteristics for verbs and adjectives are publicly available from\\\url{http://www.ims.uni-stuttgart.de/data/mid-scale}.}}
\label{sec:app-plots-croissant-noun}

\begin{figure*}[h!]
  \centering
    \includegraphics[width=\linewidth]{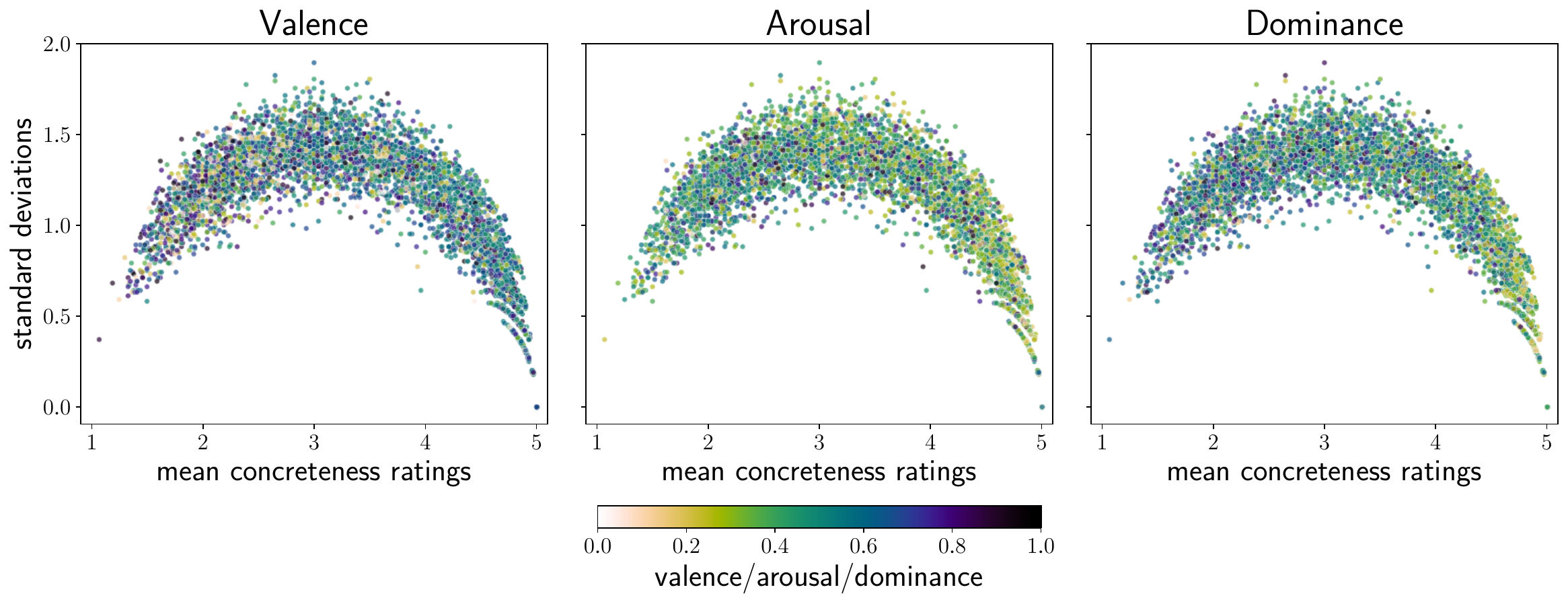}
    \caption{Mean noun ratings and standard deviations overlaid with the respective VAD scores.}
    \label{fig:vad_nouns}
  \vspace{-3mm}
\end{figure*}

\begin{figure*}[h!]
    \centering
    \includegraphics[width=.35\linewidth]{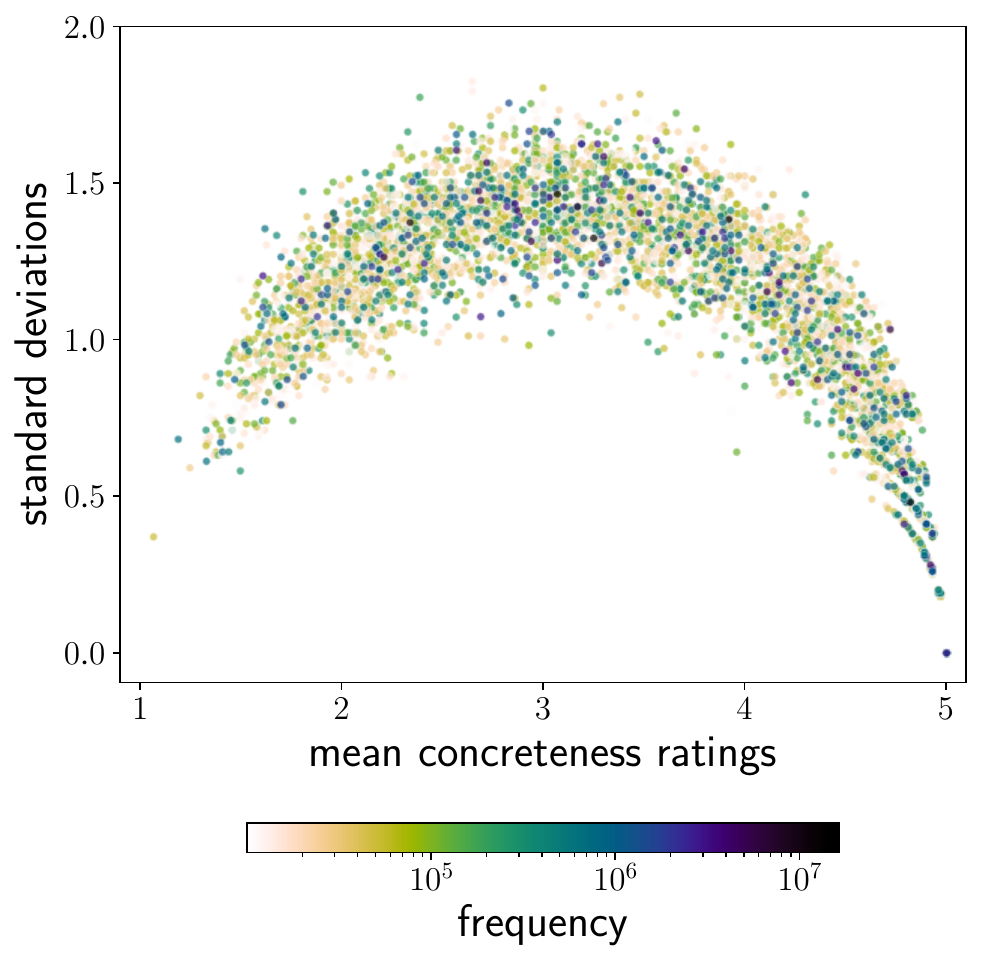}
    \includegraphics[width=.35\linewidth]{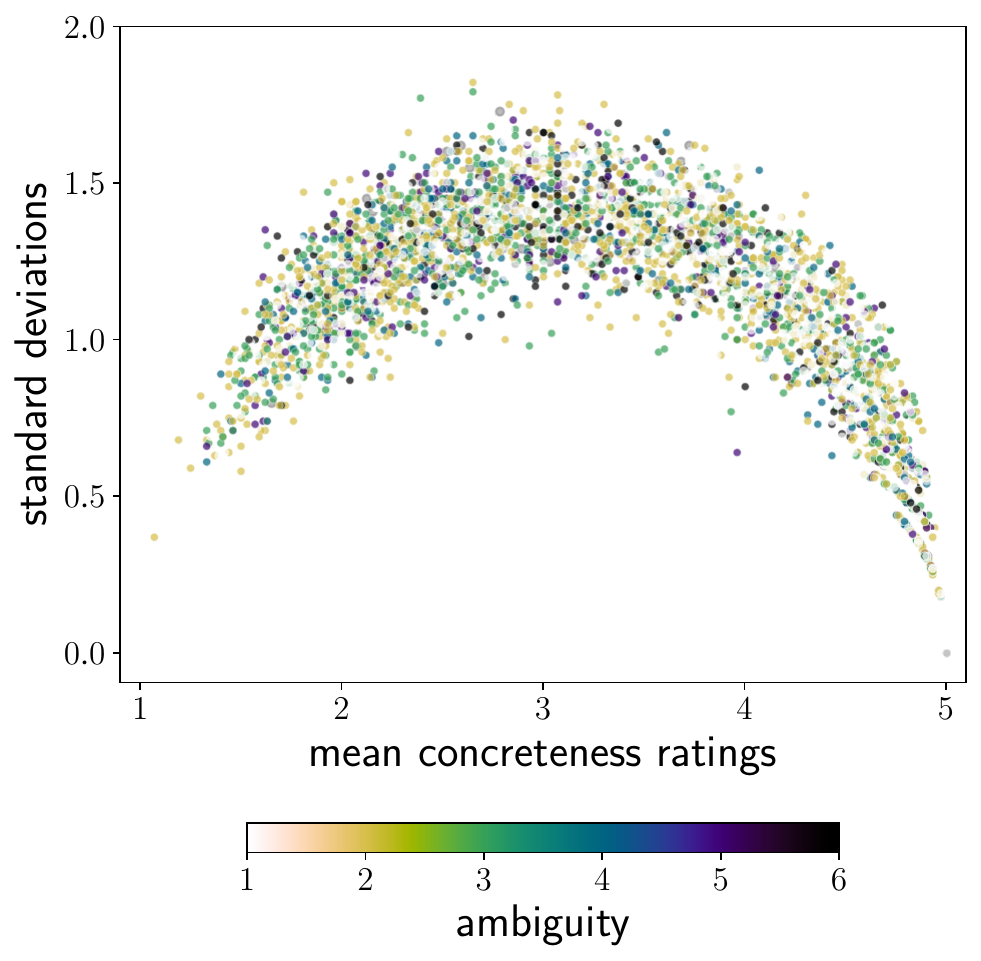}
    \caption{Mean noun ratings and standard deviations overlaid with heatmaps of the respective log$_{10}$-scaled frequency and ambiguity values.}
    \label{fig:frequency_ambiguity_nouns}
  \vspace{-3mm}
\end{figure*}

\begin{figure*}[h!]
  \centering
    \includegraphics[width=\linewidth]{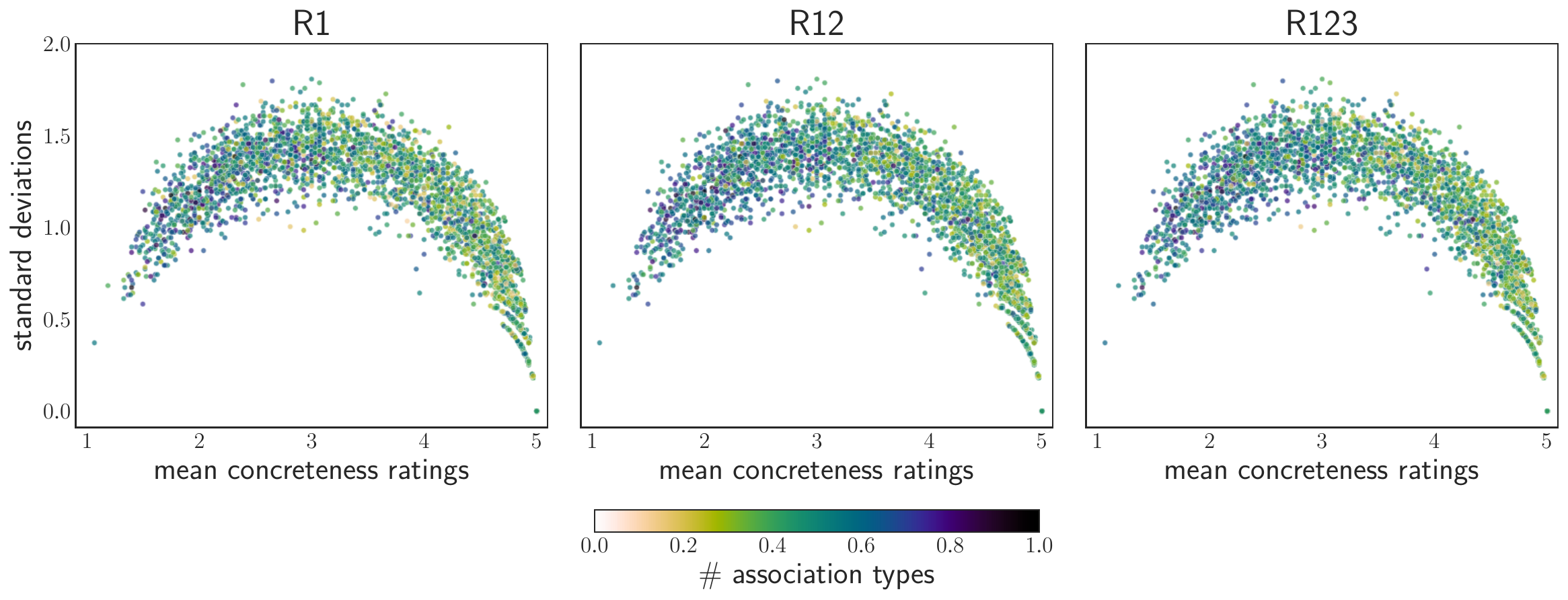}
    \caption{Mean noun ratings and standard deviations overlaid with a normalised number of the association types in the sets \textsc{R1}, \textsc{R12}, and \textsc{R123}.}
    \label{fig:word_association_nouns}
\end{figure*}

\clearpage
\section{Correlations between Target Characteristics and Concreteness: Verbs and Adjectives}
\label{sec:app-correlation-verb-adj}

\begin{table}[ht!]
\centering
\vspace{+8mm}
\begin{tabular}{llr}
\toprule

\multicolumn{2}{c}{Target characteristics} & \multicolumn{1}{c}{$\rho$} \\
\midrule

\multirow{5}{*}{Sense perception} & Auditory & -0.28$^{*}$ \\
                 & Gustatory & -0.09$^{*}$ \\
                 & Haptic & 0.47$^{*}$ \\
                 & Olfactory & 0.01$^{\textcolor{white}{*}}$ \\
                 & Visual & 0.47$^{*}$ \\
\midrule
\multirow{3}{*}{Emotion} & Valence & -0.11$^{*}$ \\
        & Affect & 0.04$^{\textcolor{white}{*}}$ \\
        & Dominance & -0.15$^{*}$ \\
\midrule
\multirow{2}{*}{Lexicon}          & Frequency & -0.01$^{\textcolor{white}{*}}$ \\
                 & Ambiguity & 0.13$^{*}$ \\
\midrule
\multirow{3}{*}{Diversity: associations} & R1 & -0.30$^{*}$ \\
                        & R12 & -0.31$^{*}$ \\
                        & R123 & -0.31$^{*}$ \\

\bottomrule
\end{tabular}

\caption{Spearman's rank-order correlation coefficient $\rho$ for the statistical relationships between degrees of concreteness and strengths of target \textit{verb} characteristics; significance level is $p<0.05$.}
\label{tab:correlationsV}
\vspace{+8mm}
\end{table}

\begin{table}[ht!]
\centering
\vspace{+8mm}
\begin{tabular}{llr}
\toprule

\multicolumn{2}{c}{Target characteristics} & \multicolumn{1}{c}{$\rho$} \\
\midrule

\multirow{5}{*}{Sense perception} & Auditory & -0.37$^{*}$ \\
                 & Gustatory & -0.01$^{\textcolor{white}{*}}$ \\
                 & Haptic & 0.35$^{*}$ \\
                 & Olfactory & 0.04$^{\textcolor{white}{*}}$ \\
                 & Visual & 0.39$^{*}$ \\
\midrule
\multirow{3}{*}{Emotion} & Valence & -0.03$^{\textcolor{white}{*}}$ \\
        & Affect & -0.07$^{*}$ \\
        & Dominance & -0.08$^{*}$ \\
\midrule
\multirow{1}{*}{Lexicon}          & Frequency & -0.04$^{\textcolor{white}{*}}$ \\
\midrule
\multirow{3}{*}{Diversity: associations} & R1 & -0.28$^{*}$ \\
                        & R12 & -0.32$^{*}$ \\
                        & R123 & -0.31$^{*}$ \\

\bottomrule
\end{tabular}

\caption{Spearman's rank-order correlation coefficient $\rho$ for the statistical relationships between degrees of concreteness and strengths of target \textit{adjective} characteristics; significance level is $p<0.05$.}
\label{tab:correlationsA}
\vspace{+8mm}
\end{table}

\clearpage
\onecolumn
\section{Mid-Scale Definitions, Ranges and Classifications across Word Classes}
\label{sec:app-mid-scale-def-results}

Intuitively, the interpretation of mid-scale targets refers to somewhere in the middle of the mean concreteness ratings plots that we have presented in Figure~\ref{fig:conc-rating-distributions}, in contrast to extremely abstract targets on the left and extremely concrete targets on the right. Accordingly, we suggest three ways of capturing this intuition, given that the number of targets per part-of-speech (POS) and also the ranges of ratings and their skewness differ across POS. We created three sets of 500 mid-scale noun targets accordingly, and also three sets of 200 mid-scale verb and 200 mid-scale adjective targets.

\begin{description}
\item[Mid-Scale-Mean] The mid-scale score is defined as the mean value on the rating scale, which is $3$ in our scale $[1; 5]$. Mid-scale targets are then defined as those words whose mean ratings are closest to $3$.
\item[Mid-Scale-Median] Given that the rating distributions differ across POS and with regard to their left vs.\ right skews, the mid-scale score is defined as the median, in our case: $3.54$ for the nouns, $2.47$ for the verbs, and $2.19$ for the adjectives. Mid-scale targets are then defined as those words whose mean ratings are closest to these medians.
\item[Mid-Scale-Median-SD] Incorporating disagreement between raters, we refine the mid-scale-median taking into account as mid-scale targets only those words whose mean ratings are closest to the median \underline{and} whose standard deviations are $>1.4$.
\end{description}

\begin{figure*}[ht!]
    \centering
    \includegraphics[width=\linewidth]{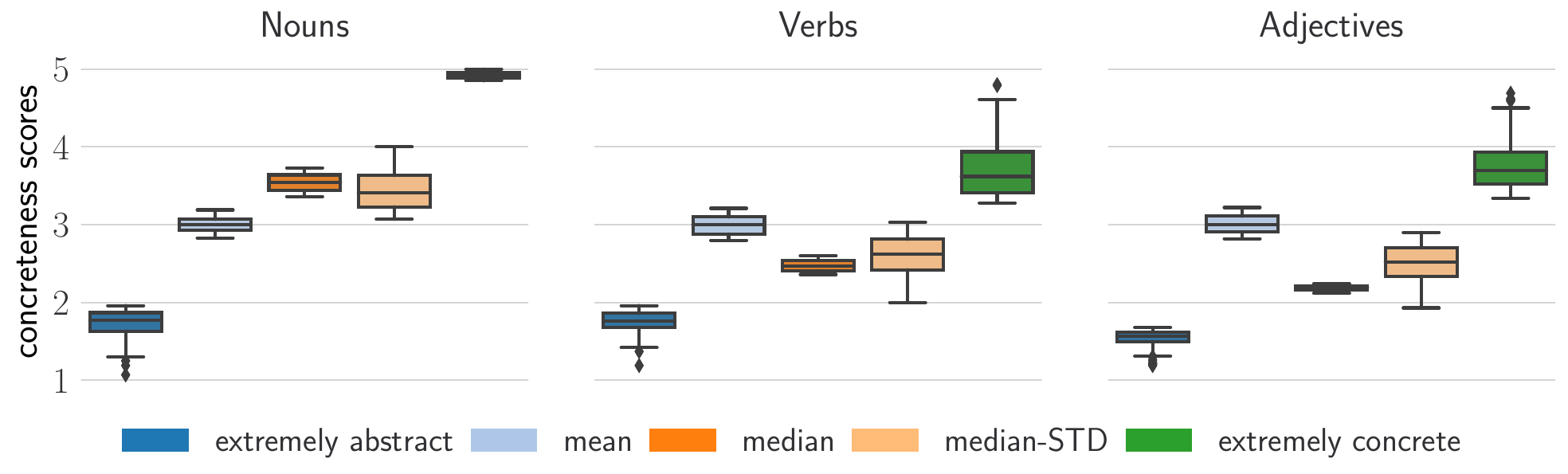}
    \caption{Distributions of concreteness scores across mid-scale definitions and POS.}
    \label{fig:mid-scale-ranges}
\end{figure*}

\begin{figure*}[ht!]
    \centering
    \includegraphics[width=\linewidth]{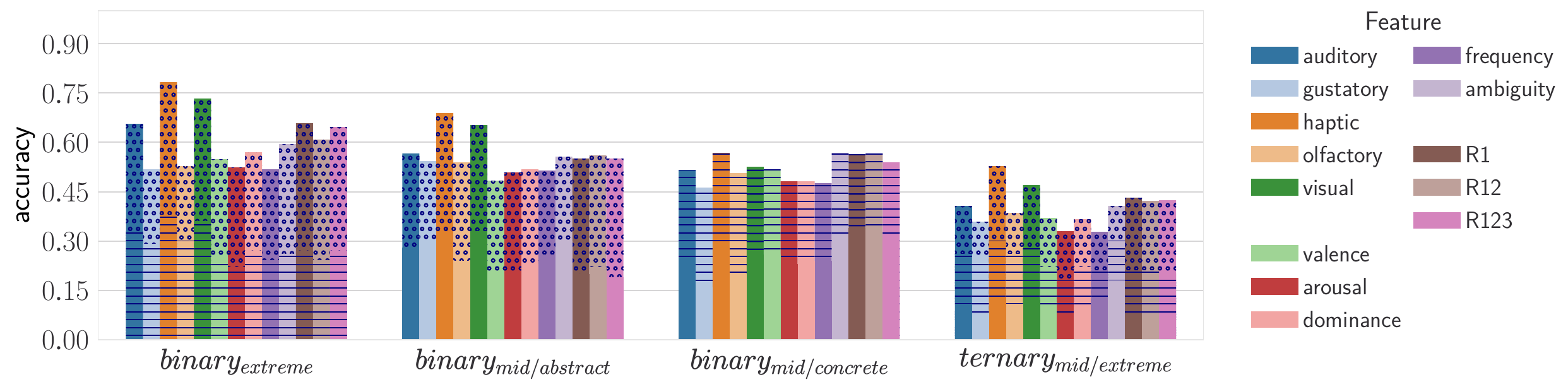}
    \caption{Results of classifications across characteristics and mid-scale/extreme experiments.
    The dotted and horizontal line patterns indicate the amount of abstract and concrete \textit{verbs} correctly classified.}
    \label{fig:mid-scale-mean-classification-results-verbs}
\end{figure*}

\begin{figure*}[ht!]
    \centering
    \includegraphics[width=\linewidth]{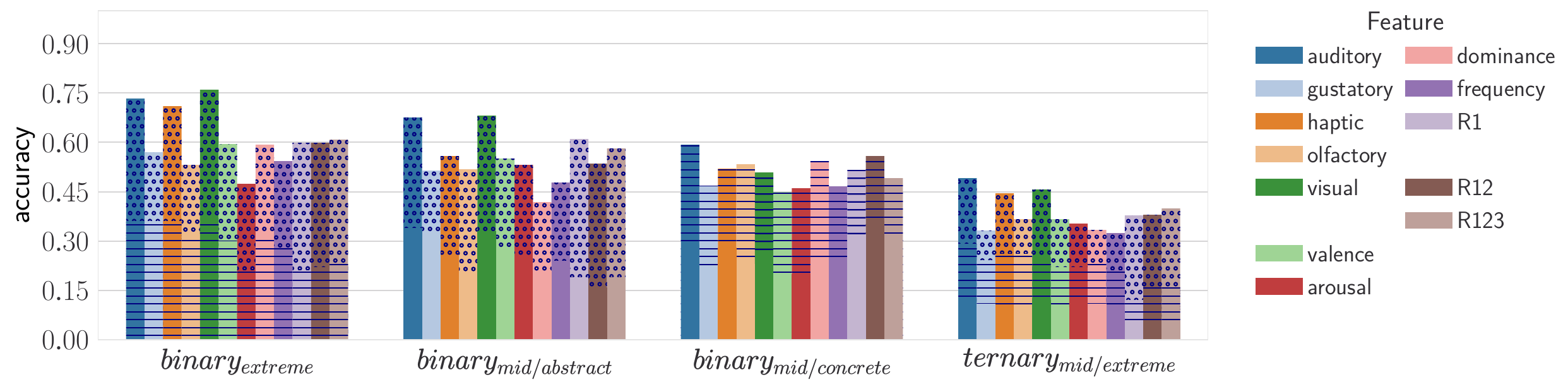}
    \caption{Results of classifications across characteristics and mid-scale/extreme experiments.
    The dotted and horizontal line patterns indicate the amount of abstract and concrete \textit{adjectives} correctly classified.}
    \label{fig:mid-scale-mean-classification-results-adjectives}
\end{figure*}

In all three cases, we selected an equal number of targets with mean ratings above and below the respective mid-scale score. Figure~\ref{fig:mid-scale-ranges} provides the mean-rating ranges of our mid-scale targets across these three mid-scale definitions, based on the respective 500/200/200 mid-scale noun/verb/adjective targets. The same figure shows the mean-rating ranges of the extremely concrete and extremely abstract targets, relying again on sets of 500/200/200 targets. We can see that the mid-scale ranges clearly differ across definitions and POS. Table~\ref{tab:mid-scale-classification-results} shows the classification results (accuracy) across these mid-scale definitions, word classes and target set constellations. Figures~\ref{fig:mid-scale-mean-classification-results-verbs} and~\ref{fig:mid-scale-mean-classification-results-adjectives} zoom into the classification results of verb/adjective targets per feature type and for the mid-scale mean definition, as done for nouns in Figure~\ref{fig:mid-scale-mean-classification-results-nouns}.

\vspace{2cm}

\vspace{2cm}

\begin{table*}[ht!]
    \centering

    \begin{tabular}{ll|>{\raggedleft\arraybackslash}p{1.5cm}|>{\raggedleft\arraybackslash}p{1.5cm}|>{\raggedleft\arraybackslash}p{1.5cm}}
    \toprule
    &                                                 & \multicolumn{3}{c}{\textbf{Mid-Scale Definition}} \\
    &
    &
    \multicolumn{1}{c}{\textbf{Mean}} &
    \multicolumn{1}{|c}{\textbf{Median}} &
    \multicolumn{1}{|c}{\textbf{Median-SD}} \\
    \midrule

    \multirow{4}{*}{nouns} 
    & \textit{binary$_{extremes}$} & 0.98 & 0.98 & 0.98 \\
    & \textit{ternary$_{mid/extremes}$} & 0.79 & 0.82 & 0.82 \\
    & \textit{binary$_{mid/concrete}$} & 0.93 & 0.91 & 0.91 \\
    & \textit{binary$_{mid/abstract}$} & 0.75 & 0.83 & 0.82 \\
    \midrule
    
    \multirow{4}{*}{verbs}
    & \textit{binary$_{extremes}$} & 0.90 & 0.90 & 0.90 \\
    & \textit{ternary$_{mid/extremes}$} & 0.63 & 0.64 & 0.65 \\
    & \textit{binary$_{mid/concrete}$} & 0.64 & 0.78 & 0.78 \\
    & \textit{binary$_{mid/abstract}$} & 0.81 & 0.65 & 0.73 \\
    \midrule
    
    \multirow{4}{*}{adjectives}
    & \textit{binary$_{extremes}$}  & 0.94 & 0.94 & 0.94 \\
    & \textit{ternary$_{mid/extremes}$}  & 0.67 & 0.67 & 0.67 \\
    & \textit{binary$_{mid/concrete}$}  & 0.68 & 0.86 & 0.81 \\
    & \textit{binary$_{mid/abstract}$}  & 0.84 & 0.55 & 0.71 \\
    \bottomrule

    \end{tabular}
    \caption{Results of the classifications across mid-scale definitions and target set constellations.}
    \label{tab:mid-scale-classification-results}
    \vspace{+10mm}
\end{table*}

\clearpage
\onecolumn
\section{Mid-Scale Disagreement Patterns in Verb and Adjective Rating Distributions}
\label{sec:app-clusters-verb-adj}

\begin{figure*}[ht!]
    \vspace{+10mm}
    \centering
    \includegraphics[width=.8\linewidth]{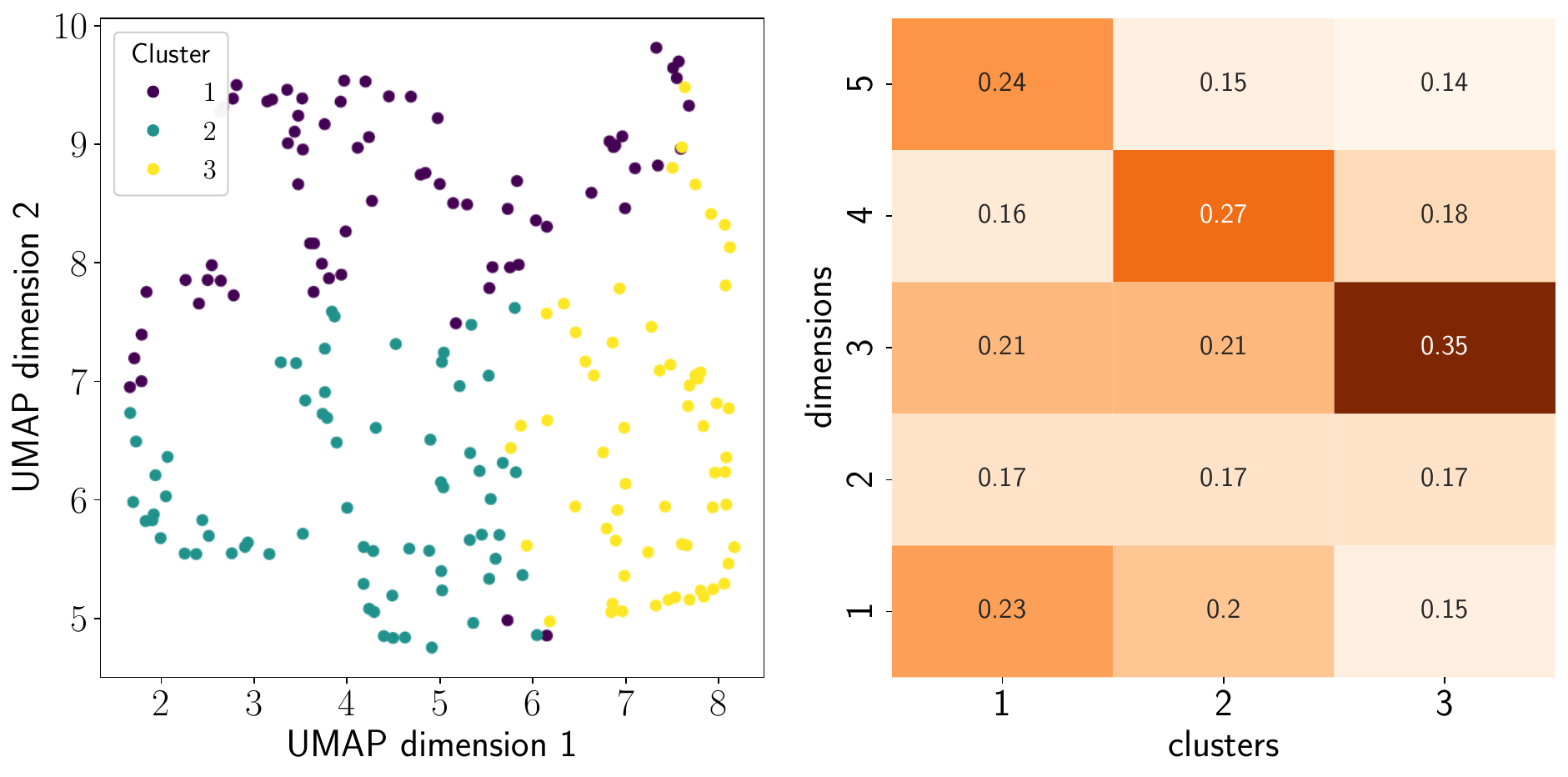}
    \caption{$k$-Means clustering ($k=3$) of 200 mid-scale verbs based on original individual per-participant rating distributions. Cluster sizes are 71, 68, and 61. The heatmap shows the rating distributions of the centroid vectors.}
    \label{fig:clustering_mean_verbs}
    \vspace{+8mm}
\end{figure*}

\begin{figure*}[ht!]
    \centering
    \includegraphics[width=.8\linewidth]{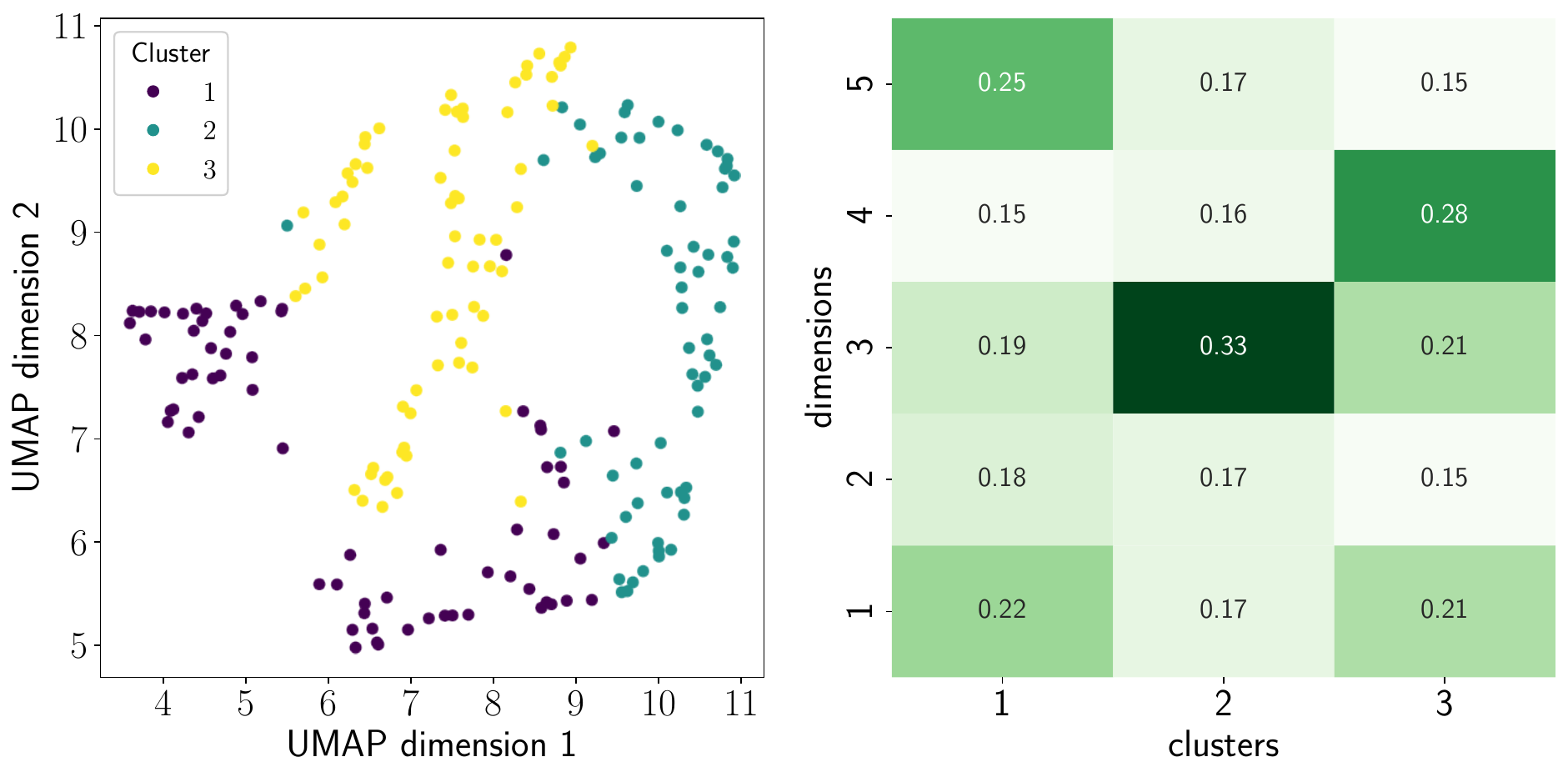}
    \caption{$k$-Means clustering ($k=3$) of 200 mid-scale adjectives based on original individual per-participant rating distributions. Cluster sizes are 68, 62, and 70. The heatmap shows the rating distributions of the centroid vectors.}
    \label{fig:clustering_mean_adjectives}
\end{figure*}

Figures~\ref{fig:clustering_mean_verbs} and~\ref{fig:clustering_mean_adjectives} present the clusters and the heat maps of rating distributions of the cluster centroids for verbs and adjectives. The clusters are based on the same $k$-Means clustering setup as those for nouns in Section~\ref{sec:analyses-disagreement}.

\end{document}